%% file: conformal.tex
\documentclass[11pt]{article}

\usepackage{amsmath,amssymb,amsfonts,mathtools}
\usepackage[margin=1in]{geometry} \usepackage{microtype} \usepackage{hyperref}
\usepackage{subcaption} \usepackage{float}

\input{macros3}

\input{local-definitions}
\input{theorenv}

\hypersetup{colorlinks=true,linkcolor=blue,urlcolor=blue,citecolor=blue}

\date{}

\begin{document}
\title{Conformal point and the calibrating conic}
\author{Richard Hartley\\
Australian National University}

\maketitle

\begin{abstract}
This gives some information about the conformal point and the calibrating conic,
and their relationship one to the other. These concepts are useful for
visualizing image geometry, and lead to intuitive ways to compute geometry, such
as angles and directions in an image.
\end{abstract}

\section{Introduction}
This paper describes two geometric objects, the calibrating conic and the
conformal point that may be overlaid on an image to aid intuitive understanding
of the geometry.  Between them, they give a visual indication of the calibration
of the camera, much more intuitive than the usual way of describing the camera
calibration in terms of an upper-triangular $3\times 3$ matrix, the calibration
matrix, and yet containing the same information in a geometric way.

The calibrating conic gives the calibration of the camera, and along with the
conformal point it gives an easy visual way of measuring angles between rays in
the image, the field of view of the camera, its orientation with respect to
major geometric features in the image, or between vanishing lines in the plane. 
It also gives a conceptually simple way, involving only geometric constructs
without algebra, of performing self-odometry under planar motion.

Both of these two constructs have been described previously in the literature,
but are perhaps unfamiliar to many readers. My purpose in revisiting them now is
as follows.

The calibration of a camera is important for making measurements in an image. 
Expressed in the traditional way, involving matrices and algebraic constructs it
is, however, a little mysterious.  Without practice, one cannot look directly at
an image and immediately guess the camera calibration. Neither can one display
the calibration matrix in a document and expect the reader to fully grasp its
connection with a given image. It is the purpose of this note to show how the
calibrating conic, and its partner, the conformal point with respect to lines in
the image, for instance the horizon, give a direct cue from which one
immediately understands such basics as the field of view of the camera or its
tilt with respect to a plane, and is able to make easy computation of angles via
a simple geometric construct.

It will be demonstrated that these geometric objects are often easily visible
and derivable from an image, without algebra or optimization. This direct
geometric interpretation of the image geometry is relevant in an era of deep
learning.  If one accepts a hypothesis that neural networks are better at
performing tasks that the human visual system can easily do, rather than
accurately compute matrices, homographies, absolute conics, trifocal tensors or
the mathematical paraphernalia of multiview geometry, then it makes sense to
look at visual and intuitive ways of presenting the same sort of data. It is
beyond the scope of this note to demonstrate the use of concepts such as these
in a learning context; the paper confines itself to an demonstration of their
properties.

Part of the approach elaborated in this paper belongs to the tradition of
elementary geometry, in that many of the computations are carried out by
ruler-and-compass constructions.  It is a pleasant conceit that this aspect of
the work would have appealed to Euclid and his school.

\subsection{Motivation and prior work}
The motivation for preparing this document was a talk I was asked to give at a
workshop at CVPR-2025 on the topic of calibration and pose.  My particular
assignment was to talk about classical methods. In pondering what I could talk
about that was not entirely known to members of the audience knowledgeable in
multiview geometry, I hit upon the idea of talking about two concepts that have
been largely ignored in the literature.  The first is the calibrating conic,
which although mentioned in the Hartley-Zisserman book has not been taken up by
the wider community.  The second concept is that of the conformal point, which
was introduced in a paper from 2003: ``Visual Navigation in the Plane using the
Conformal Point,'' \cite{conformal-point} which has received $7$ citations in
the succeeding $22$ years, a respectable tally by most expectations.

Since the conformal point paper is not easily accessible online, at least not
through my library, it seemed opportune to describe its properties and its
connection to the calibrating conic.

\subsection{Basics}

The calibrating conic was introduced in the Hartley-Zisserman book
\cite{HartleyZisserman:03} as an alternative to the image of the absolute conic
(IAC) to express the calibration of a camera.

The IAC is a conic lying in the image plane represented by the symmetric matrix
$\omega = (\m K \m K\tr)^{-1}$, where $\m K$ is the calibration matrix of
the camera, a $3\times 3$ upper triangular matrix.%
\footnote{ The conic represented by a matrix $\omega$ is the set of points $\v
x$ in the plane (expressed in homogeneous coordinates) satisfying $\v x\tr
\omega \v x = 0$.
}
It is useful as a theoretical tool for investigating the camera calibration, and
to deduce properties of the scene.

Notable among the properties of the IAC is that it allows the computation of the
angle between two rays represented by points in the image. Thus, given two rays
passing through the camera centre, corresponding to points $\v x_1$ and $\v x_2$
in the image, the angle between the two rays is given by
\begin{equation}
\label{eq:IAC-angle-formula}
\cos (\theta) = \frac {\v x_1\tr \omega \v x_2}{ \sqrt{\v x_1\tr \omega \v x_1} \, \sqrt{\v x_2\tr \omega \v x_2}} ~.
\end{equation}
This is equation (8.9) in \cite{HartleyZisserman:03}%
\footnote{ Equation numbers in \cite{HartleyZisserman:03} refer to the second
edition.
}
In particular, if the two rays are orthogonal, then
\begin{equation}
\label{eq:IAC-orthogonal-vanishing-condition}
\v x_1 \tr \omega \v x_2 = 0 ~.
\end{equation}
\paragraph{Vanishing points. }
These considerations correspond most particularly to vanishing points in the
image.  These are the points in the image corresponding to particular directions
in space, such as the vertical direction, or either of two horizontal direction.
In particular, \eq{IAC-angle-formula} applies in the case that $\v x_1$ and $\v
x_2$ are vanishing points in two directions.  In particular, for orthogonal
vanishing points (for instance the vertical and horizontal vanishing points,
\eq{IAC-orthogonal-vanishing-condition} holds.

\paragraph{Lines on a plane. }
The IAC may also be used to determine the angle between two lines lying on a
plane in space.  For this, it is also necessary to identify the vanishing line
of the plane (in the image).  This is called the {\em horizon line} for the
plane, denoted $\v l_\infty$.  In such a case, the vanishing point of the two
lines lying on the plane is obtained by extending the lines to the horizon. The
point of intersection of the two lines with the horizon gives the points in the
image corresponding to the vanishing direction of the two lines. The angle
between these two lines is equal to the angle between their vanishing
directions.

\paragraph{Invariance. } A basic property of the IAC is that it is invariant both
to translation and rotation of the camera. Chapter 8 (particularly Section 8.5)
of \cite{HartleyZisserman:03} gives further details of the use of the IAC in
camera calibration, and shows how certain restrictions on the calibration matrix
(such as zero-skew, known principal point) can be used to facilitate
calibration.  A popular calibration technique of Zhang
(\cite{ZY-Zhang-calibration} essentially leverages the invariance property of
the IAC under homographies of images of a planar calibration object captured by
a rotating and translating camera.

\paragraph{Invisibility. }
Although the IAC may be conceived as a conic, represented by a $3\times 3$
symmetric matrix, it is a purely imaginary conic, that contains no real points. 
There is no point $\v x$, in homogeneous coordinates that satisfies the equation
$\v x\tr \omega \v x = 0$. For this reason, all computations with the IAC need
to be carried out algebraically, and one's intuition regarding this conic is
limited.  One cannot point to a real conic delineated in the image and identify
it as the IAC.  For this reason, \cite{HartleyZisserman:03} in section 8.5
introduce the
{\em calibrating conic}, a real conic that can be seen in the image, and that
shares some of the useful properties of the IAC.


\section{The Calibrating Conic}
\label{sec:calibrating-conic}
\begin{figure}[h]
\centerline{
\includegraphics[width=5.5in]{"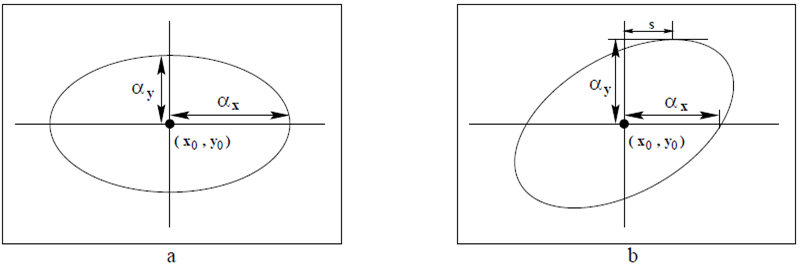"}
}
\caption{For images with square pixels, the calibrating conic is a circle with centre at the principal
point of the image.  If the pixels are rectangular with non-square aspect ration, then the
calibrating conic is an axis-aligned ellipse.  Finally if skew is present, the conic is an
non-axis-aligned ellipse from which one can read the skew parameter as shown.
}
\label{fig:reading-parameters}
\end{figure}

The {\em calibrating conic} introduced in section 8.10, page 231 of 
\cite{HartleyZisserman:03} is the image of all points that lie at an angle of
$45$ degrees from the optical axis. For a camera with calibration matrix $\m K$,
the calibrating conic is represented by the matrix
\begin{equation}
\label{eq:calibrating conic}
\m C = \m K\mtr \mbegin{ccc} 1 & & \\ & 1 & \\ && -1 \mend \m K^{-1} 
\end{equation}
It is shown in \cite{HartleyZisserman:03}, figure 8.27, reproduced here as
\fig{reading-parameters}
%
%
how the entries of the calibration matrix $\m K$ can be read directly and
visually from the calibrating conic.  Thus, the calibrating conic gives a direct
geometric visual represention of the camera calibration.  This is in contrast to
the IAC or even the calibration matrix itself which give a more abstract
representation of the calibration.

If the calibrating conic is known, then it may be drawn directly on top of the
image itself, allowing an enhanced understanding of the geometry of the scene. 
This is shown in the illustrative images to follow.

As shown in\cite{HartleyZisserman:03}, page 231 the calibrating conic can be
written as
\[
\m C = (\m K\mtr \m K^{-1}) (\m K \m D \m K^{-1}) = \omega \m S
\]
where $\m D$ is the matrix $\diag(1, 1, -1)$ appearing in \eq{calibrating
conic}.  Moreover $\m S = \m K \m D \m K^{-1}$ represents reflection of a point
through the centre of the calibrating conic. Following
\cite{HartleyZisserman:03} one denotes $\m S \v x$, for a point $\v x$ by
$\dot{\v x}$.  Then it follows (\cite{HartleyZisserman:03} (8.19)) that
\begin{equation}
\v x'{}\tr \omega \v x = \v x'{}\tr \m C \dot{\v x} ~.
\end{equation}

\paragraph{Polar with respect to a conic. }
If $\m C$ is a $3\times 3$ matrix representing a conic, and $\v x$ is a point,
then $\m C \v x$ represents a line, known as the {\em polar} of the point $\v x$
with respect to the conic.  Geometrically, the line $\m C \v x$ is constructed
as follows.

If $\v x$ lies exterior to the conic, let $\v y_1$ and $\v y_2$ be two point on
the conic such that the lines $\v x \v y_1$ and $\v x \v y_2$ are tangent to $\m
C$.  The polar of $\v x$ is the line passing through $\v y_1$ and $\v y_2$.

\begin{figure}[H]
\centerline{
\includegraphics[height=2.0in]{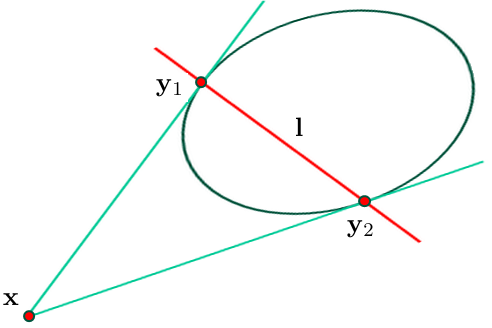} \hskip 0.25in
\includegraphics[height=2.0in]{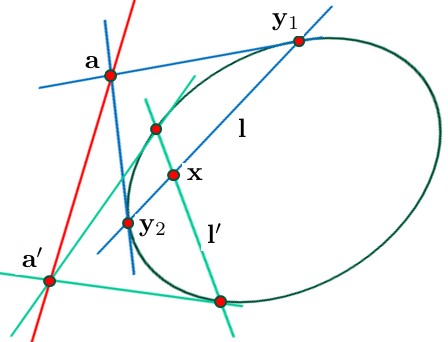} 
}
\caption{Construction of the polar of a point $\v x$ with respect to a conic.  The red line
is the polar of point $\v x$, as described in the text.}
\label{fig:polar-construction}
\end{figure}

If $\v x$ lies in the interior of the conic then this construction is not
possible (except algebraically).  In this case, let $\v l$ be a line passing
through $\v x$, meeting the conic at points $\v y_1$ and $\v y_2$.  Construct
tangents to $\m C$ at $\v y_1$ and $\v y_2$, meeting at a point $\v a$.  Do this
again for a different line $\v l'$, resulting in a different point $\v a'$.  The
polar of $\v x$ is the line joining $\v a$ and $\v a'$.  These two constructions
are illustrated in \fig{polar-construction}.

\begin{figure}[h]
\centerline{
\includegraphics[height=2.25in]{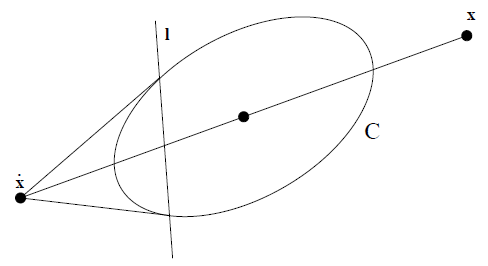}
}
\caption{\em(Diagram taken from \cite{HartleyZisserman:03} figure 8.28).
Finding the directions perpendicular to a given direction.  Given point $\v x$, the line $\v l$
is the reflected polar of $\v x$ with respect to the calbrating conic $\m C$. Any point on the line $\v l$ represents
a direction perpendicular to $\v x$.
}
\label{fig:reflected-polar}
\end{figure}

With this interpretation, $\omega \v x = \m C \m S \v x = \m C \dot{\v x}$ is
the polar of $\dot{\vs x}$ with respect to the calibrating conic.  This line may
be termed the {\em reflected polar} of $\v x$ (with respect to the calibrating
conic).  Thus, the condition $\v x'\tr \omega \v x = 0$ means that the point $\v
x'$ lies on the reflected polar $\m C \dot{\v x}$, or by symmetry that $\v x$
lies on the reflected polar $\m C \dot{\v x}'$. Since the condition $\v x'\tr
\omega \v x = 0$ means that $\v x$ and $\v x'$ represent orthogonal vanishing
directions the following theorem is proved.
\begin{theorem}
Rays represented by points $\v x$ and $\v x'$ are orthogonal if and only if $\v
x'$ lies on the reflected polar $\m C \dot{\v x}$.
\end{theorem}
This is essentially what is stated by \cite{HartleyZisserman:03} Result 8.30,
and illustrated by \fig{reflected-polar} (which is figure 8.28 of
\cite{HartleyZisserman:03});
%

\paragraph{Measuring angles using the calibrated conic. }
Equation \eq{IAC-angle-formula} gives a way to compute the angle between any two
rays.  However, it is an algebraic formula only and cannot be easily interpreted
intuitively or geometrically. It will be shown next, however, that the angle
between two rays may be computed via a geometric construction using the
calibrating conic.

\begin{figure}[h]
\centerline{
\includegraphics[height=2.0in]{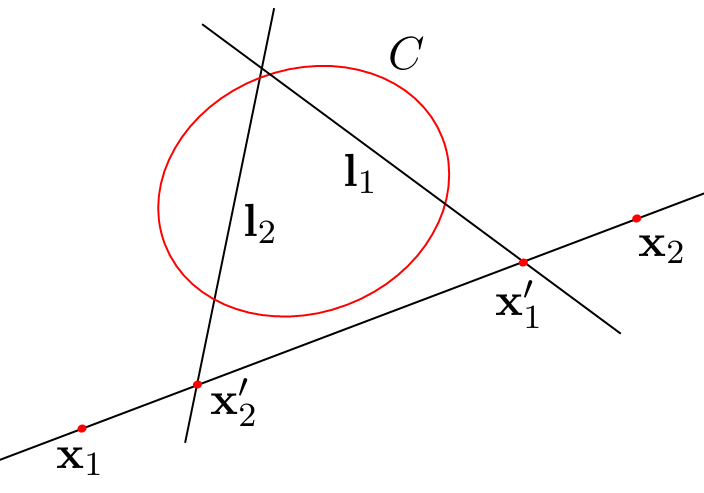}
}
\caption{The construction to compute the angle $\theta$ between two points $\v x_1$ and $\v x_2$, using
the calibrating conic.  Lines $\v l_1$ and $\v l_2$ are the reflected polars of $\v x_1$ and $\v x_2$.  
They intersect the line joining $\v x_1$ and $\v x_2$ at points $\v x_1'$ and $\v x_2'$ respectively.
Then $\cos^2(\theta)$ is given by the cross ratio \eq{angle-formula}.  In addition, $\cos(\theta)$ is
negative, because $\v x_1$ and $\v x_2$ lie on opposite sides of $\v l_1$.
}
\label{fig:cross-ratio-construction}
\end{figure}

\begin{theorem}
\label{thm:angle-formula}
Let $\v x_1$ and $\v x_2$ be two points and $\m C$ the calibrating conic. Denote
by $\v l_1 = \m C \dot{\v x}_1$ and $\v l_2 = \m C \dot{\v x}_2$, their
reflected polars. Construct the line through $\v x_1$ and $\v x_2$; call it $\v
x_1 \times \v x_2$. Let $\v x_1'$ be the intersection of $\v x_1 \times \v x_2$
with $\v l_1$, and $\v x_2'$ the intersection of $\v x_1 \times \v x_2$ with $\v
l_2$. Then
\begin{equation}
\label{eq:angle-formula}
\cos^2 (\theta) = \frac{|\v x_2 \v x'_1| \, |\v x_1 \v x'_2|} 
{|\v x_1 \v x'_1| \, |\v x_2 \v x'_2|} ~,
\end{equation}
where $|\v x_i \v x'_j|$ represents the (Euclidean) distance between the points.

Moreover, $\cos(\theta)$ is positive or negative depending on whether both $\v
x_1$ and $\v x_2$ lie on the same side or opposite sides of $\v l_1$ (or equally
well, $\v l_2$) respectively.
\end{theorem}
Note that this expression is the cross-ratio of collinear points $\v x_1, \v
x_2, \v x_1', \v x_2'$. This construction is illustrated in
\fig{cross-ratio-construction}.

\begin{proof}
We start with the equation \eq{IAC-angle-formula} and rewrite it as
\begin{align*}
\cos (\theta) &= \frac {\v x_1\tr \m C \,\dot{\v x}_2}
{ \sqrt{\v x_1\tr \m C\, \dot{\v x}_1} \, \sqrt{\v x_2\tr \m C \,\dot{\v x}_2}} \\[0.1in]
&=\frac {\v x_1\tr \v l_2}
{ \sqrt{\v x_1\tr \v l_1} \, \sqrt{\v x_2\tr \v l_2} }
\end{align*}
By symmetry of $\omega$, the numerator can also be written as $\v x_2\tr \m C
\,\dot{\v x}_1$, so
\begin{align*}
\cos (\theta) 
&=\frac {\v x_2\tr \v l_1}
{ \sqrt{\v x_1\tr \v l_1} \, \sqrt{\v x_2\tr \v l_2} }
\end{align*}
Multiplying both forms of this equation together, results in
\begin{align*}
\cos^2 (\theta)
&=\frac {(\v x_1\tr \v l_2 ) \,(\v x_2\tr \v l_1 )}
{ (\v x_1\tr \v l_1 ) \, (\v x_2\tr \v l_2 ) } ~.
\end{align*}

Apart from appriate normalizing factor, each of $\v x_i\tr \v l_j$ represents
the perpendicular distance of the point $\v x_i$ to the line $\v l_j$.  However,
the normalizing factors cancel top and bottom, so we obtain
\begin{align*}
\cos^2 (\theta)
&=\frac {d(\v x_1, \v l_2 )\,d(\v x_2, \v l_1 ) } 
{d(\v x_1, \v l_1 )\,d(\v x_2, \v l_2 ) } \\[0.1in]
&=\frac {d(\v x_2, \v l_1 )}
{ d(\v x_1, \v l_1 ) } 
~\times~
\frac {d(\v x_1, \v l_2 )}
{  d(\v x_2, \v l_2 ) }
~.
\end{align*}
Now, these distances represent perpendicular distances.  However, by similar
triangles,
\[
\frac {d(\v x_2, \v l_1 )} { d(\v x_1, \v l_1 ) } = 
\frac {d(\v x_2, \v x_1' )} { d(\v x_1, \v x_1') }   =
\frac {|\v x_2 \v x_1' |} { |\v x_1 \v x_1'| } 
\] 
where $\v x_1'$ is the point of intersection of the line $\v x_1 \times \v x_2$
with line $\v l_1$; siimilarly for the other ratio. Putting this all together
results in
\[
\cos^2 (\theta) = \frac{|\v x_2 \v x'_1| \, |\v x_1 \v x'_2|} 
{|\v x_1 \v x'_1| \, |\v x_2 \v x'_2|} ~,
\]
as the required cross-ratio.

This formula alone does not determine whether $\cos(\theta)$ is positive or
negative, or equivalently, whether $\theta$ is less than or greater than
$90^\circ$.  This ambiguity is resolved by whether $\v x_1$ and $\v x_2$ lie on
the same side of $\v l_1$ (or $\v l_2$), or not. Proof of this condition is left
as an exercise for readers with time on their hands to fill out a rainy Sunday.
\end{proof}


\begin{figure}[H]
\centerline{
\includegraphics[width=6.5in]{"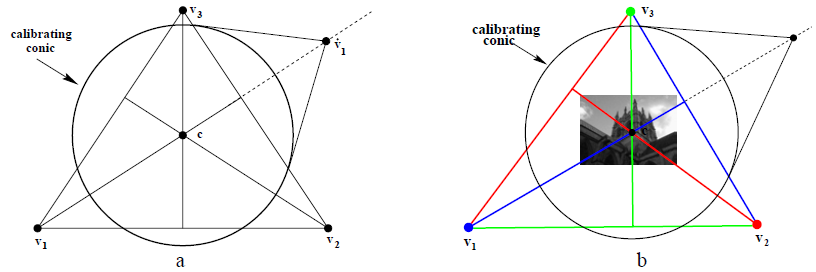"}
}
\caption{Finding the calibrating conic given three orthogonal vanishing points.  
This diagram is taken directly from \cite{HartleyZisserman:03}.  It shows how the calibrating conic
is at the orthocentre of the triangle formed by three vanishing points, $\v v_1$, $\v v_2$ and $\v v_3$.
The radius is defined by the condition that the reflected polar of (say) $\v v_1$ passes through $\v v_2$ and $\v v_3$.
}
\label{fig:three-orthogonal}
\end{figure}


\section{The conformal point}
The previous discussion derives an algebraic formula involving the cross-ratio
for $\cos^2(\theta)$, where $\theta$ is the angle between two rays. As nice as
this formula is, it is still not very intuitive, being a mixture of algebra,
trigonometry and geometry.  The angle $\theta$ is not easily visualized, given
the two points $\v x_1$, $\v x_2$ and the calibrating conic $\m C$.  In the
present section a purely geometric method is given for constructing the angle
between two rays.  This method involves a point called the ``conformal point'',
which is defined with respect to the vanishing line in the image of a world
plane (the horizon line), or with respect to a line joining two points $A$ and
$B$ in an image.

In the discussion of conformal point, it is assumed always that the pixels have
square aspect ratio.

\begin{figure}[H]
\centerline{
\includegraphics[height=2.0in]{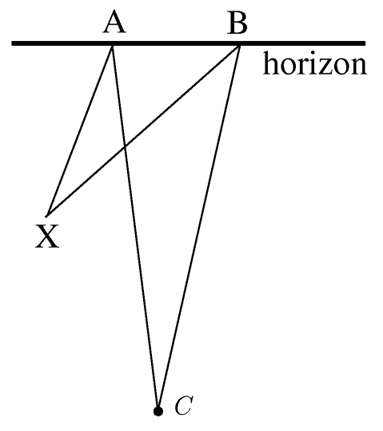}
}
\caption{Measuring angles between rays using the conformal point.  Let two lines lie on a world
plane.  Let the projections of these lines meet at a point $X$ in an image.
Assuming that the camera is viewing the plane obliquely, the angle $\widehat X$ is not equal to
the true angle as measured in the world.  To compute the true {\em world angle}, in the image extend the two
lines to points $A$ and $B$ on the vanishing line of the plane, known as the horizon.  Then
connect these two points back to the {\em conformal point} $C$, which is a point to be described
presently.  The angle $\widehat{ACB}$ is equal to the true angle in the world plane between the
lines.  This construction works for any two lines in the world plane.
}
\label{fig:not-referenced-1}
\end{figure}

The conformal point was introduced in the paper \cite{conformal-point} presented
at the IRSS conference (International Symposium on Robotics Research), in Lorne
Australia in 2001.  However, the complete paper does not seem to be available
easily on line
For this reason, its main results and derivations are repeated here.

The derivation of the conformal point and its construction is described in the
subsequent images.


\begin{figure}[H]
\centerline{
\includegraphics[height=3.5in]{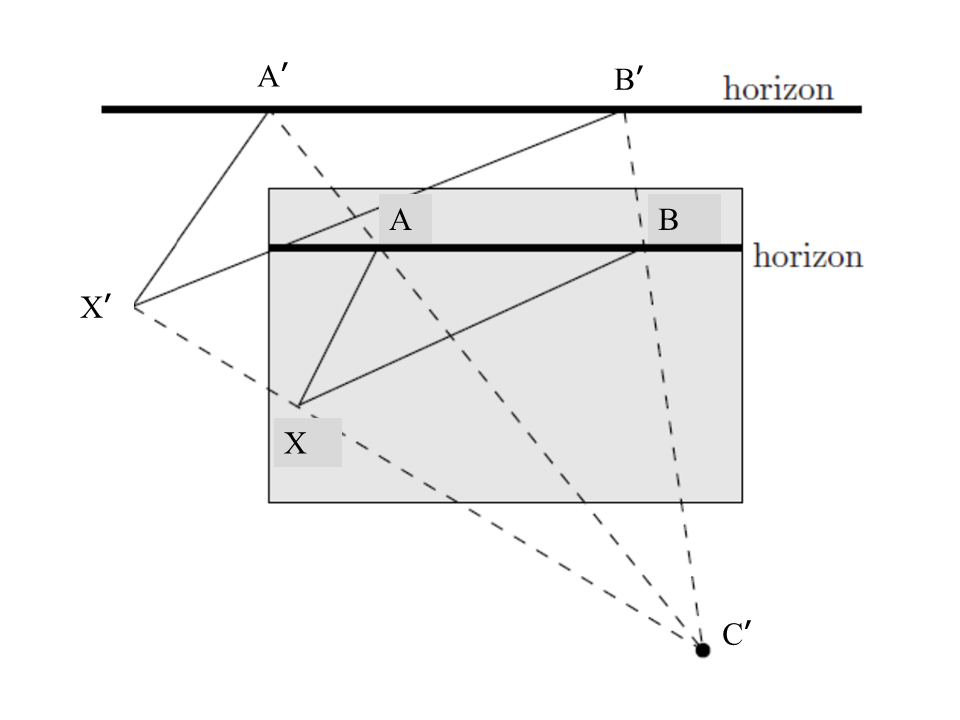}
}
\caption{Finding the angle between lines on a plane.  In this figure, letters with a prime ($A'$, $B'$, $C'$ and $X'$)
represent points in the world, whereas unprimed letters ($A$, $B$, $X$) represent points in the image plane, the projections
of the corresponding world points.  The point $C'$ is the camera centre, or centre of projection, $X'$ is the point where two lines in a world
plane meet, and $A'$ and $B'$ are the vanishing points of these lines, that is the points where they meet the
vanishing line of the plane, otherwise known as the horizon or line at infinity in the plane. \\
Since lines $X'A'$ and $C'A'$ meet at a point at infinity, $A'$, they are parallel, written $X'A' \parallel C'A'$.  Similarly
$X'B' \parallel C'B'$.  It follows that
$\widehat{X'} = \widehat {A' X' B'} = \widehat{A'C'B'} = \widehat{AC'B}$.\\
In other words, the angle between two lines is equal to the angle between the rays from the camera centre
to the vanishing points in the image of the two lines.  This will be interpreted as an angle measured in the image, in the
next figures.
}
\label{fig:original-proof}
\end{figure}
\begin{figure}
\centerline{
\includegraphics[height=2.3in]{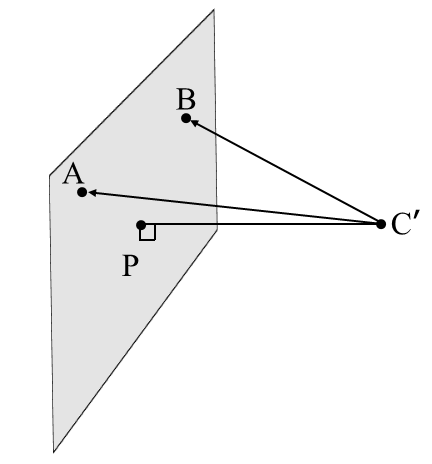}
}
\caption{Following on from \fig{original-proof}, the task is to find the angle between rays $C'A$ and $C'B$.  Point $C'$ is the camera
centre; $P$ is the principal point, namely the
foot of the perpendicular from $C'$ to the image plane.  
\\ 
The task is to determine the angle between the rays corresponding
to pixels $A$ and $B$, that is the angle $\widehat{A C' B}$, which was shown in
\fig{original-proof} to equal the angle (in the world) between to lines
vanishing at points $A'$ and $B'$.
}
\label{fig:proof-1}
\end{figure}

\begin{figure}
\centerline{
\includegraphics[height=2.3in]{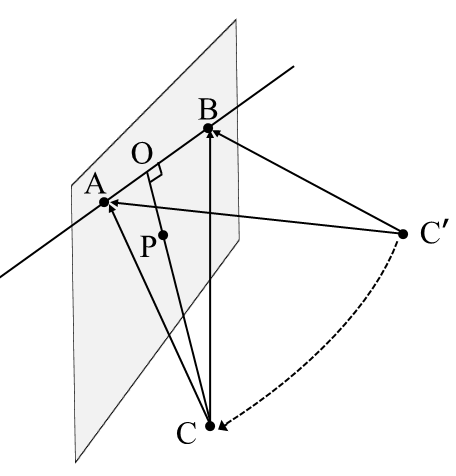}
}
\caption{The angle $\widehat{AC'B}$ is related to an angle $\widehat{ACB}$ measured directly
in the image.  The point $C$ is constructed by rotating the triangle $AC'B$ about the line $AB$ until
the camera centre $C'$ is rotated onto the image plane, resulting in the point $C$.
Thus, the angle $\widehat{ACB}$ is equal to angle $\widehat{AC'B}$, the angle between the two rays
corresponding to image points $A$ and $B$.  \\
The point $C$ so constructed does not depend explicitly on the two points $A$ and $B$, but only
on the line through them.  Consequently, the angle between rays corresponding to any two points
in the line $AB$ can be measured by the angle subtended by connecting them to point $C$.  This
point is referred to as the conformal point corresponding to the line $AB$.\\
The position of the conformal point constructed in this way lies along the perpendicular to the
line $AB$ passing through the principal point $P$.  Its distance from $O$ is equal to the distance
$|OC'|$.  The length $|PC'|$ is the focal length $f$, and let $|OP| = d$. The triangle $OPC'$ is a right-angled
triangle (since $P$ is the principal point), so $|OC| = |OC'| = \sqrt{f^2 + d^2}$.\\
There are in fact two conformal points, on either side and equally distant from the horizon.  In
the following, we shall without comment choose one of these.
}
\label{fig:lines-on-plane-proof}|
\end{figure}

\begin{figure}
\centerline{
\includegraphics[height=2.8in]{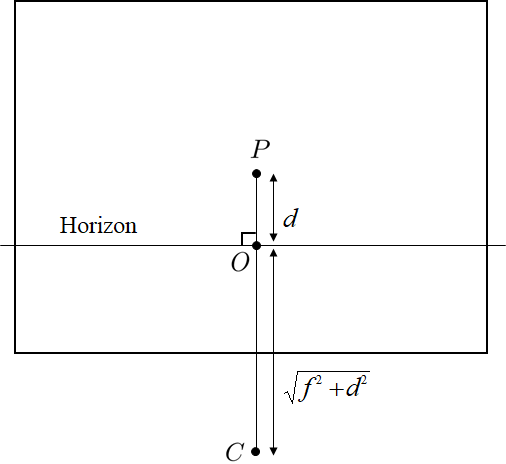}\hskip 0.5in
\includegraphics[height=2.8in]{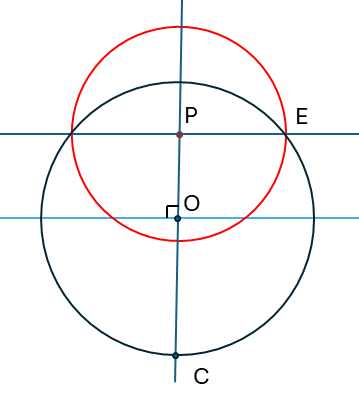}
}
\caption{The conformal point in an image relative to a plane 
is the point on a perpendicular to the horizon
(vanishing line of the plane) at a distance $\sqrt{f^2 + d^2}$ from the 
horizon; $d$ is the distance of the horizon from the principal point,
and $f$ is the focal length. \\
The image on the right shows a geometric construction that computes the conformal point
given the calibrating conic (red) and the horizon line (light blue). Find point $E$ where the 
line parallel to the horizon meets the calibrating conic, and draw a circle with centre $O$
and radius $|OE| = \sqrt{f^2 + d^2}$.  This circle meets the line $PO$, perpendicular to the horizon,
at the conformal point $C$.\\
Conversely, if the conformal point $C$, principal point $P$ and horizon are known, then the 
calibrating conic can be constructed as the circle with centre $P$ and radius equal to
$PE$, where $E$ is the intersection of the circle with centre $O$ passing through $C$, and the
line through $P$ parallel with the horizon.\\
There are two conformal points, one above and one below the horizon.  
One of them will always lie inside the calibrating conic
and one will lie outside.
}
\label{fig:conformal-geometry}
\end{figure}


\clearpage

\clearpage
\subsection{Why is it called ``conformal point''? A Riemannian perspective }

This section is not essential for understanding the rest of this note, but gives
an interpretation of the conformal point in the context of Riemannian geometry
and the theory of conformal mappings.

In complex analysis or Riemannian Geometry, a conformal mapping is one that
preserves angles.
We consider a function $f:A\rightarrow B$ where $B$ is a metric space with
metric $d_B$, and $f$ is injective. In this case we can define a metric $d_A$ on
$A$ by
\[
d_A(\v x, \v y) := d_B\big(f(\v x), f(\v y)\big) ~.
\]
It is straightforward to check that $d_A$ satisfies the axioms of a metric,
since $d_B$ is a metric and $f$ is injective.

Now we specialize to the case where $A$ and $B$ are open subsets of
$\mathbb{R}^2$, and $f$ is at least differentiable. The {\em  differential} of
$f$ at $\v x$,  is a linear map
\[
df_{\v x} : T_{\v x}A \longrightarrow T_{f(\v x)}B,
\]
the \emph{push-forward} of tangent vector. Here, for each $\v x\in A$ and $\v
y\in B$, we denote the tangent spaces by $T_{\v x}A$ and $T_{\v y}B$,
respectively; these can be identified with $\mathbb{R}^2$. We equip $A$ and $B$
with Riemannian metrics $g$ and $h$, that is, for each point we have inner
products
\[
g_{\v x} : T_{\v x}A \times T_{\v x}A \to \mathbb{R},
\qquad
h_{\v y} : T_{\v y}B \times T_{\v y}B \to \mathbb{R}.
\]
(In the simplest case, $g$ and $h$ are just the usual Euclidean inner products
on $\mathbb{R}^2$.)

We assume that the differential map $df_{\v x}$ has full rank everywhere on $A$.
In this case, we can define a new inner product on $T_{\v x}A$, the
\emph{pull-back} of $h$ by $f$, as
\[
(f^*h)_{\v x}(\v v,\v w)
\;:=\;
h_{f(\v x)}\!\big(df_{\v x}(\v v),\,df_{\v x}(\v w)\big) \text{~~ for~~}
\v v,\v w \in T_{\v x}A.
\]
This $f^*h$ is again a Riemannian metric on $A$, provided $df_{\v x}$ has full
rank.

We say that $f$ is \emph{conformal at} $\v x$ (with respect to $g$ and $h$) if
there exists a scalar $\lambda(\v x)>0$ such that
\[
(f^*h)_{\v x}(\v v,\v w)
\;=\;
\lambda(\v x)\, g_{\v x}(\v v,\v w)
\quad\text{for all } \v v,\v w\in T_{\v x}A,
\]
in other words, the pull-back metric at $\v x$ is equal, up to scale $\lambda(\v
x)$ to the original metric $g$ on $A$. More briefly, at $\v x$ we have $f^*h =
\lambda(\v x) g$. The consequence of this is that the angle between vectors at
$\v x$ is preserved. In particular, defining the angle via
\[
\cos \angle (\v v, \v w) = \frac{g_{\v x} (\v v, \v w)}{\sqrt{g_{\v x} (\v v, \v v)}\sqrt{g_{\v x} (\v w, \v w)}}
\]
it follows that
\[
\angle (\v v, \v w) = \angle (df_{\v x}(\v v), df_{\v x} (\v w)) 
\]
for all vectors $\v v, \v w \in T_{\v x}(A)$.

In the special case where $A,B\subset\mathbb{R}^2$ are equipped with the
standard Euclidean metrics, we may choose orthonormal bases for $T_{\v x}A$ and
$T_{f(\v x)}B$. Then the push-forward $df_{\v x}:T_{\v x}A\to T_{f(\v x)}B$ is
represented by a $2\times 2$ matrix $J_{\v x}$ (the Jacobian of $f$ at $\v x$),
and the conformality condition is equivalent to
\[
J_{\v x}^{\mathsf T} J_{\v x} \;=\; \lambda(\v x) I.
\]
Now, this may be applied to the case of a camera (assuming square pixels) views
a plane (the world plane). The inverse of the camera projection defines a
mapping from part of the image ``below'' the horizon to the part of the world
plane in front of the camera.  This is a bijective projective mapping. By the
above construction, the standard Euclidean inner product in the world plane can
by pulled back to define a metric in the image plane. This mapping is conformal
at a single point below the horizon.  Direct computation, involving computation
of the Jacobian of this mapping, shows that this point is the one previously
called the conformal point.  It is the only point at which the image-to-plane
mapping is a conformal mapping in the standard sense.  Angles are preserved at
this point.

This computation may be carried out by hand, or with computer assistance,
resulting in the idenfication, as before, that the conformal point is the point
lying on the perpendicular from the principal point to the horizon at a distance
$\sqrt{f^2 + d^2}$ from the horizon.


\subsection{Construction the calibrating conic}

\begin{figure}[h]
\centerline{
\includegraphics[width=3.5in]{"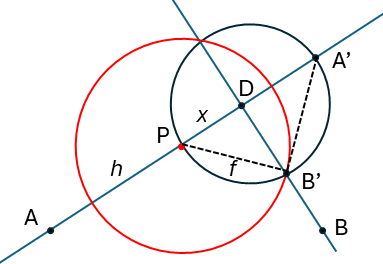"}
}
\caption{Constructing the calibrated conic given two orthogonal rays, $A$ and $B$ and
the principal point $P$, assuming square pixels. Let $A'$ be the reflection of $A$ through $P$.  
Let $D$ be
the intersection of $AA'$ with the perpendicular through $B$, and let this line
meet the circle with diameter $PA'$ (the black circle) at point $B'$.  The calibrated conic is 
the (red) circle with centre $P$ passing through $B'$. \\
According to the construction, the angle $\widehat{A'B'P}$ is a right-angle,
so $A'B'$ is tangent to the calibrated conic, and the line through B is the polar of $A'$.\\
Let $|PD| = x$ and $|PA'| = |PA| = h$ as shown.  Since triangles $A'B'P$ and $B'DP$ are similar
one derives, $x / f = f / h$, so $f^2 = xh$.
}
\label{fig:construction-f1}
\end{figure}

\begin{figure}
\centerline{
\includegraphics[width=3.5in]{"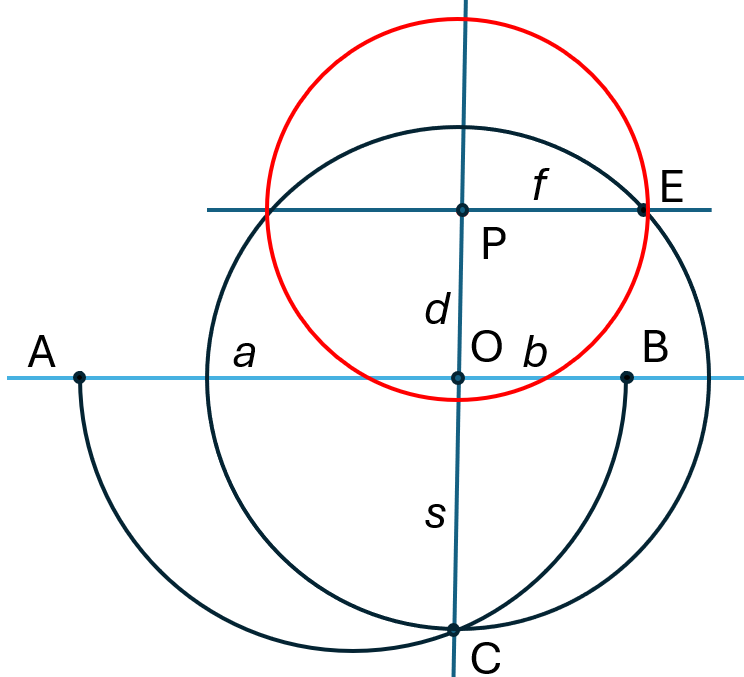"}
}
\caption{Second way to compute the calibrating conic and focal length $f$
using the conformal point.
Let $A$ and $B$ be two points in the image corresponding to orthogonal rays, and 
let $P$ the principal point (centre) of the image.
Construct a (semi-)circle with diameter $AB$, and form its intersection point,
$C$ with the perpendicular from $P$ to the line $AB$.  Point $C$
is the conformal point, since the angle $\widehat{ACB}$ is a right-angle by
construction (angle in a semi-circle).  \\
Next form a circle with centre $O$ and radius $OC$.  This meets the line through
$P$ parallel to $AB$ at point $E$.  Finally the calibrated conic has centre $P$
and radius $PE$.  \\
Denoting $s = |OC|$, 
by construction, $|OC|^2 = |AO| \, |OB|$, or $s^2 = a b$.  Moreover, $|OE|^2 = |OC|^2 = a b$,
and since $OPE$ is a right-angled triangle, it follows, defining $f = |PE|$
that $f^2 = |OE|^2 - |OP|^2 = s^2 - d^2$.  Alternatively $f^2 = ab - d^2$.
\\
The estimate of $f$ in this diagram and in \fig{construction-f2} give different ways
to estimate $f$.  It will be shown in \fig{construction-final-step} that the results are the same.
}
\label{fig:construction-f2}
\end{figure}

The following diagrams show two ways to compute the focal length -- it is
assumed that the principal point is known, and pixels are square.  Thus, the
calibrating conic is just a circle of radius $f$ centred at the principal point.
 Two ways of computing the calibrated conic from this minimal information, known
principal point, square pixels and a pair of orthogonal directions, are given in
\fig{construction-f1} and \fig{construction-f2}.

\begin{figure}
\centerline{
\includegraphics[width=2.5in]{"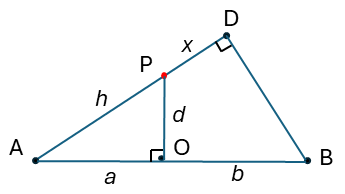"}
}
\caption{\em
Showing that the two estimates of $f$ derived in 
\fig{construction-f1} and \fig{construction-f2} are equal comes down to showing that $ab - d^2 = xh$.
This figure shows the essential geometry required to make this demonstration.
Since triangles $AOP$ and $ADB$ are similar, it follows that
$h / a = (a+b) / (h+x)$, which gives $h^2 + hx = a^2 + ab$.  But, $h^2 - a^2 = d^2$, so
$hx = ab - d^2$, as required. 
}
\label{fig:construction-final-step}
\end{figure}


\begin{figure}
\centerline{
\includegraphics[width=2.5in]{"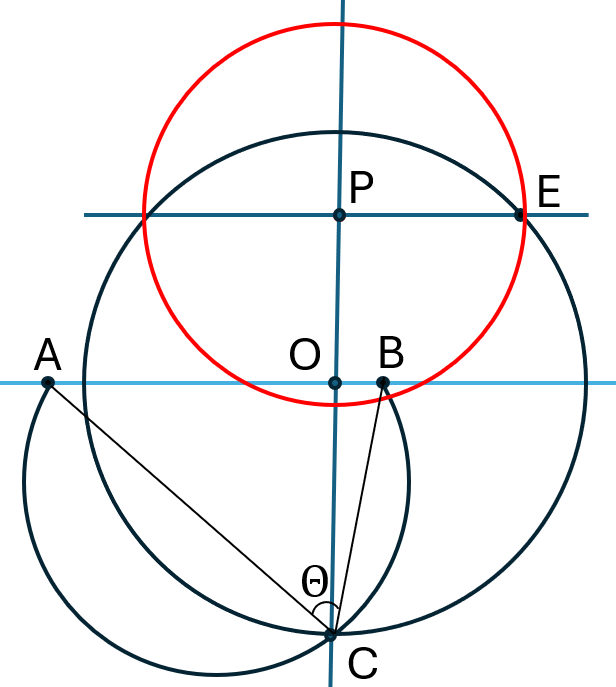"}
}
\caption{\em
The construction of the calibrating conic by the conformal point method of \fig{construction-f2}
is a little more complicated than that in \fig{construction-f1}, the reflected-conic method.  However,
it has the advantage that it can be generalized to the case of rays that are not necessarily
orthogonal but are separated by any given known angle, $\theta$.  The only modification required
is that the circle with diameter $AB$ passing through points $A$, $B$ and $C$ is replaced by
a $\theta$-angle circle passing through $A$ and $B$.  For instance, by elementary Euclidean geometry,
for any other point $C'$ lying
on the circle through $A,B,C$ angle $\widehat{AC'B}$ equals $\widehat{ACB}$.  Thus, if rays $A$ and $B$
are known to be at an angle $\theta$ and the circle is constructed with this property then it will pass
through the conformal point $C$.  (Details of how to construct an angle-$\theta$ circle are left to the
reader. )\\
The rest of the construction of the calibrating conic are identical.
}
\label{fig:construction-acute-angle}
\end{figure}
\clearpage

\subsection{Measuring angles}
Next we consider ways to measure angles between rays represented by points in
the image.  This relates particularly to vanishing points in the image
corresponding to directions in space, or to angles on a plane.

\begin{figure}[h]
\centerline{
\includegraphics[height=2.1in]{"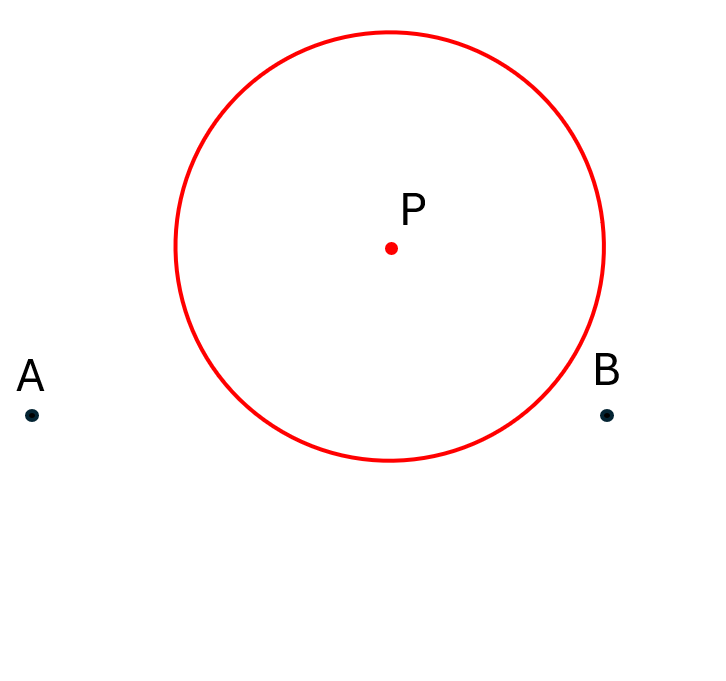"} \hskip 0.5in
\includegraphics[height=2.5in]{"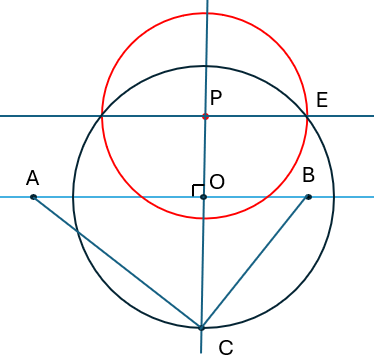"}
}
\caption{
How to measure angles between two rays $A$ and $B$ given 
only the calibrating conic $\m C$ (left).  The steps (right) are as follows:\\
Construct the line $AB$ and the perpendicular line $PO$. \\
Where is $E$ is the point on the calibrating conic along the radius parallel with $AB$, construct the circle
with centre $O$ and radius $|OE|$.  This circle meets the extension of the line $PO$ at point
$C$, which is the conformal point. \\
The angle between rays $A$ and $B$ is equal to the angle 
$\widehat{ACB}$.  \\
Moreover, the angle between any other points $A'$ and $B'$ on the same line $AB$ is given
by the angle that they subtend at the conformal point.
}
\label{fig:measuring-angle}
\end{figure}

\clearpage

\begin{figure}
\centerline{
\includegraphics[height=2.0in]{"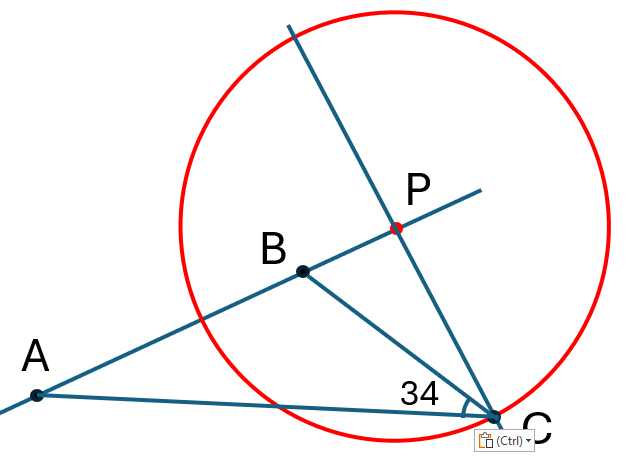"} 
}
\caption{\em
Computation of angles between rays that are in the same plane as the principal ray
is particularly easy.  In this case, the principal point $P$ and the two points $A$ and $B$ are collinear
in the image.  In this case, the distance $d=0$, (the distance from $P$ to the line $AB$) and
the conformal point is at distance $f$, hence on the calibrating conic, in a direction perpendicular to the line
$AB$.  The angle between the rays can then be measured directly; in this case it is $34^\circ$.
}
\label{fig:collinear-points}
\end{figure}
\begin{figure}
\centerline{
\includegraphics[height=2.0in]{"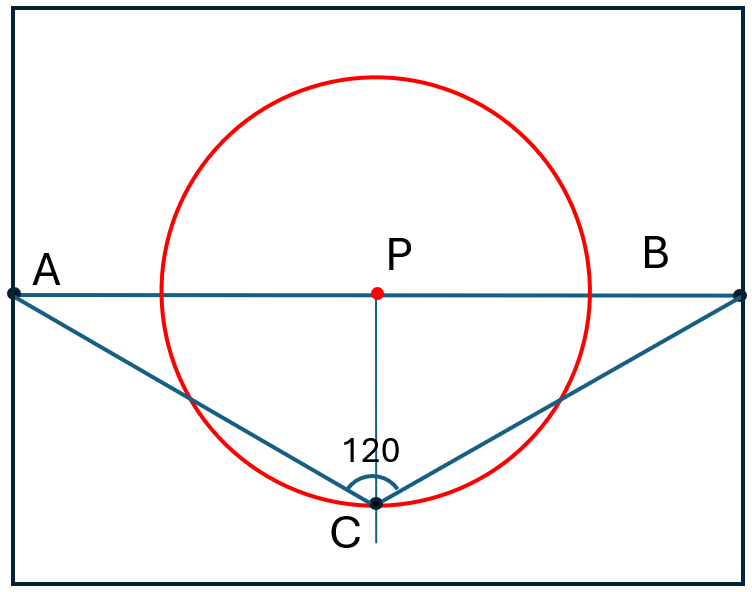"} \hskip 0.5in
\includegraphics[height=2.0in]{"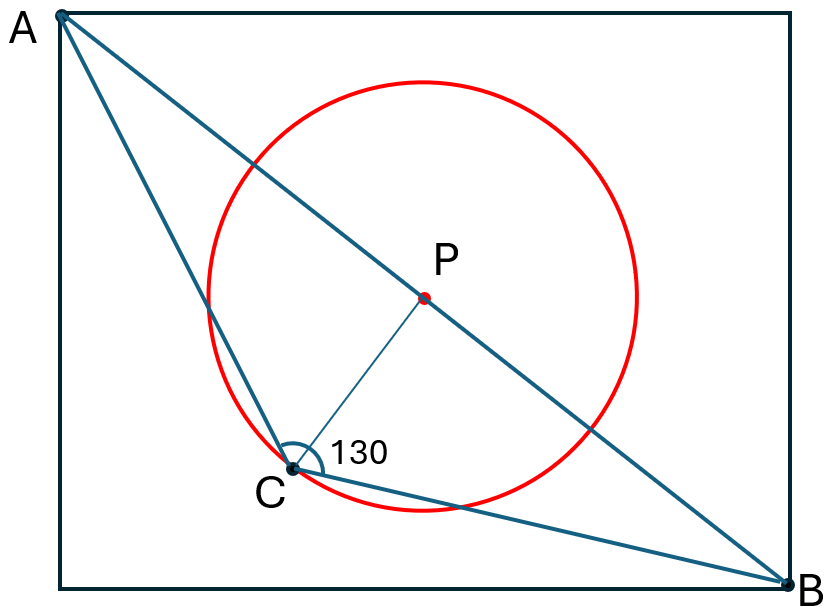"}
}
\caption{
Computing the field of view using the calibrating conic.  The method described in \fig{collinear-points}
applies directly to computing the field of view of an image.  The conformal point lines on the
calibrating conic, and the field of view is equal to the angle subtended by two points at the edges 
of the image.
Direct construction, and measurement
of angles shows that the field of view of the image (rectangular box) are $120^\circ$ horizontal
and $130^\circ$ diagonal.
$\widehat{ACB}$.
}
\label{fig:FOV}
\end{figure}

\begin{figure}
\centerline{
\includegraphics[height=3.5in]{"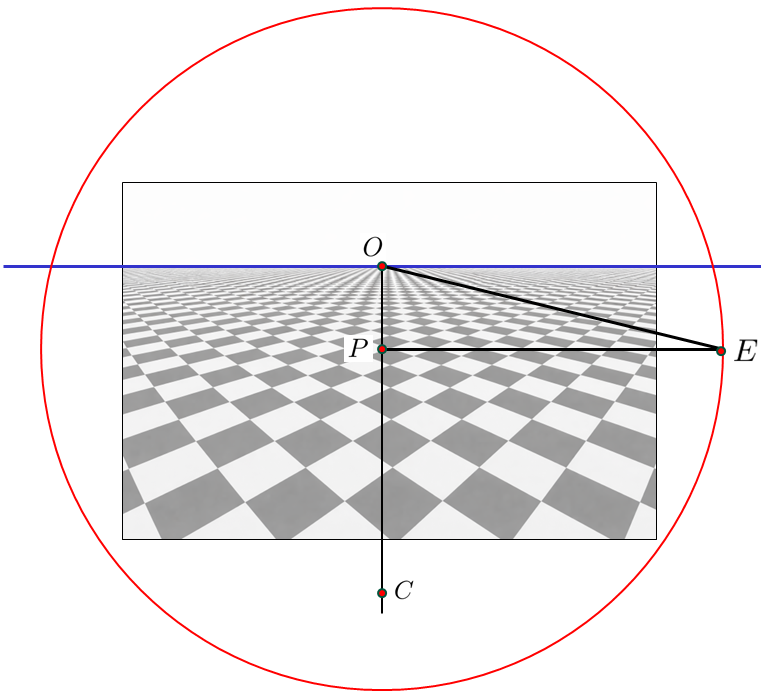"} 
}
\caption{
The calibrating conic can also be used to estimate the tilt of the camera with respect to a plane,
for instance the ground plane.  In this case, the angle to be measured is the one between
the principal ray of the camera and the horizon.  This is equal to the angle between the 
rays corresponding to the principal point $P$, and the foot of the perpendicular
to the horizon, point $O$.  Since the line $OP$ passes through the principal point, the
conformal point for this line is the point $E$, perpendicular to line $OP$, lying on the 
calibrating conic.   Therefore, the tilt of the camera is equal to the angle $\widehat{OEP}$,
roughly estimated at $13.7^\circ$ downward tilt.
\\
In this figure, $C$ is the estimated position of the conformal point for the horizon line (blue) -- that is the
point where the corners of the tiles would appear square.  This point was used to estimate the
calibrating conic, as described previously, but is not necessary if the calibrating conic is known.  
Point $E$ is the conformal point
for the line $OP$, perpendicular to the horizon. 
}
\label{fig:camera-tilt}
\end{figure}

\clearpage


\subsection{Calibration of the camera for images with 
tiled floors}

If an image shows a tiled grid, perhaps a tiled floor, then calibration is
especially easy.  In this case, vanishing points either of the grid lines, or
else two pairs of diagonals can be used to calibrate.  Assuming principal point
at the centre, and square pixels, the previous constructions can be used to find
the calibrating conic.  This is illustrated in the following figures.

\begin{figure}[h]
\centerline{
\includegraphics[width=4.5in]{"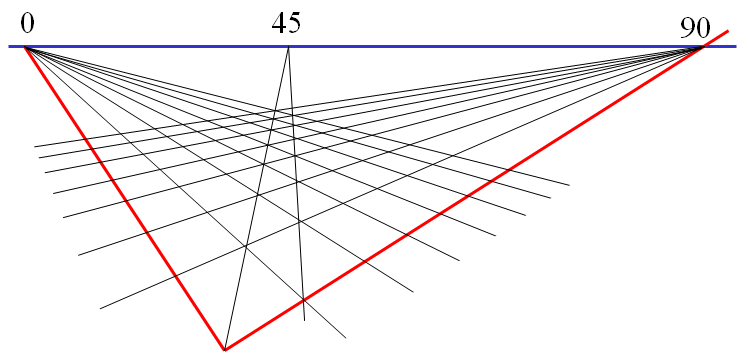"}
}
\caption{Computation of orthogonal directions as vanishing points
of orthogonal lines 
on a plane.  Assuming that checker-board markings are square or rectangular,
two orthogonal vanishing points, and the vanishing line of the plane are
easily constructed.  Diagonals also give the $45^\circ$ vanishing direction.\\
Conversely, knowledge of these three vanishing points allows one to
draw perspectively correct grids.\\
Note that knowing two angles (here two $45^\circ$ angles are visible) gives one constraint
for the location of the principal point, since the conformal point $C$ must lie on the 
intersection of two $45^\circ$ circles, and the principal point is along the
perpendicular to the horizon from $C$.
}
\label{fig:checker-board}
\end{figure}

\begin{figure}[H]
\centerline{
\includegraphics[width=7.0in]{"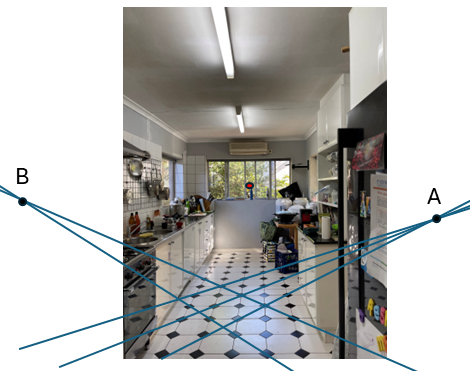"}
}
\caption{Kitchen scene with centre of image and two orthogonal
vanishing points marked.
}
\label{fig:kitchen}
\end{figure}


\begin{figure}[H]
\centerline{
\includegraphics[width=8.0in]{"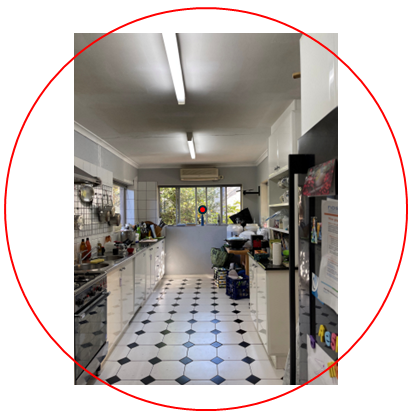"}
}
\caption{Calibrating conic, computed by the previous construction,
overlaid on the image.  To view this image with the correct perspective
(as seen from the camera position) the eye should be placed at the level
of the principal point, at a distance equal to the distance from the principal
point to the overlaid calibrating conic.
}
\label{fig:kitchen-with-conic}
\end{figure}

\clearpage

\subsection{Outdoor scenes}
Road intersections are often perpendicular.  This can allow the calibration of
the camera, as shown in the following figures.


\begin{figure}[H]
\centerline{
\includegraphics[width=7.0in]{"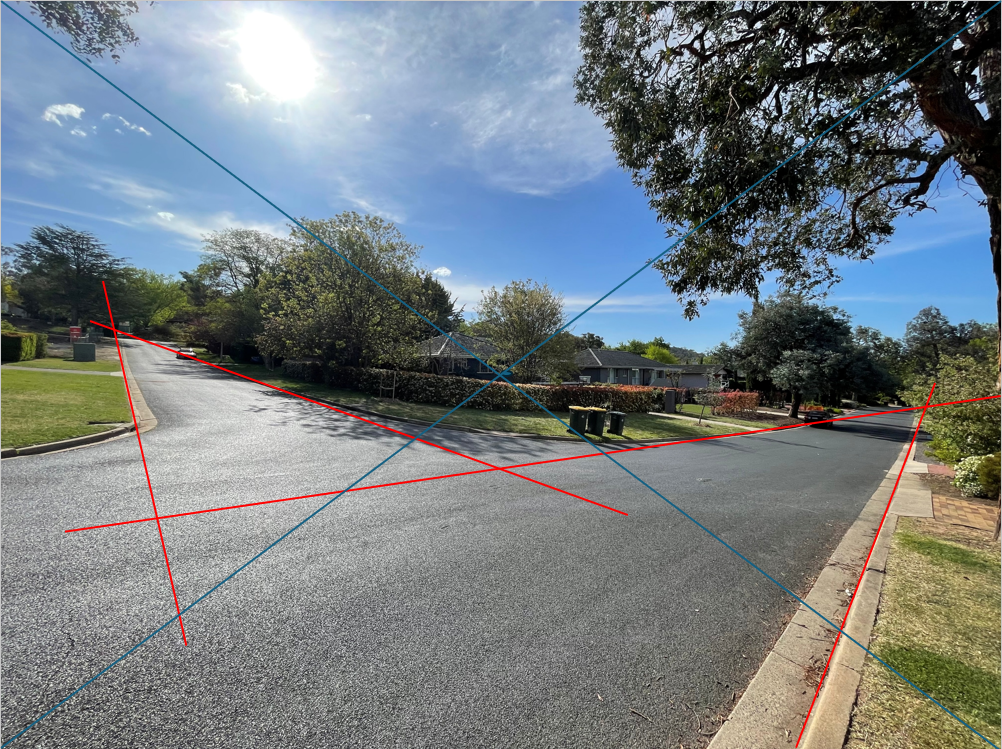"}
}
\caption{A street scene showing two roads, known to be at right-angles.  The task is to calibrate the camera 
given the usual conditions of square pixels and principal point at the centre. \\
As a first step, the horizon line and vanishing points of the two roads are constructed.  The ``horizon'' is 
slanted because the road goes up to the left and down to the right.
}
\label{fig:road-scene}
\end{figure}


\begin{figure}[H]
\centerline{
\includegraphics[width=8.0in]{"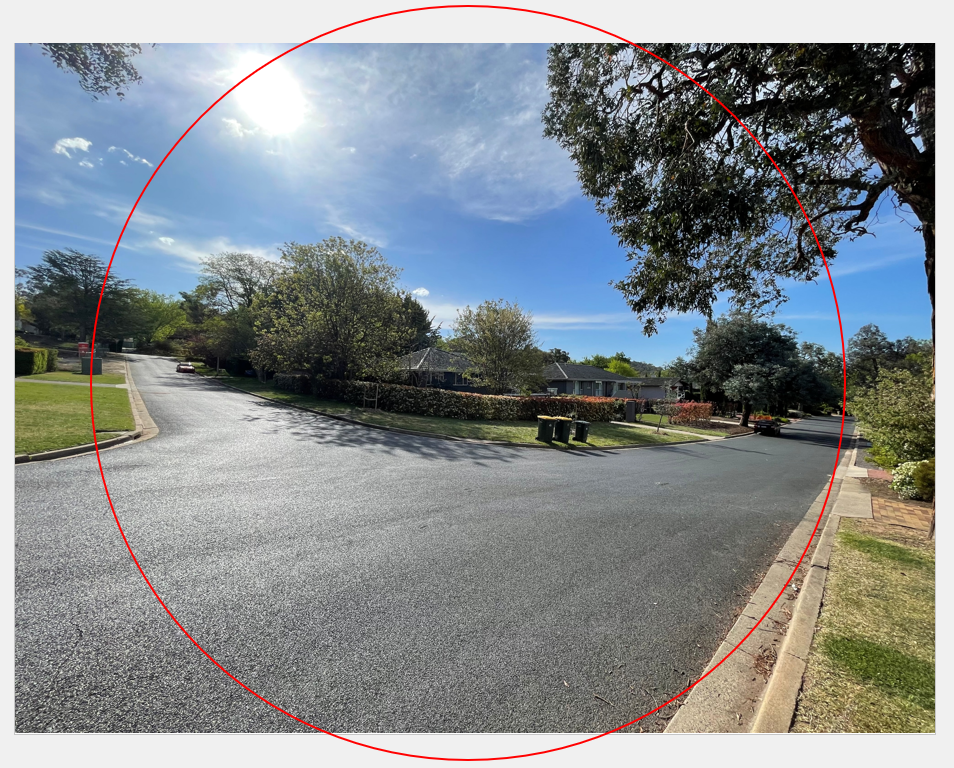"}
}
\caption{Overlaid calibrating conic, constructed using the technique given in
\fig{construction-f1}.  The field of view is now readily apparent at a little over $45$ degrees.
}
\label{fig:Slide47}
\end{figure}


\clearpage

\subsection{Calibration of paintings}
The techniques described here can also be used to calibrate the apparent camera
associated with a painting.  This technique was pioneered in the work of
Criminisi \cite{Criminisi:01,CriminisiKempZisserman:05}.  Using computer
geometry tools to analyze paintings is also treated in detail in a more recent
book by David Stork \cite{Stork:24a}.

Identifying the calibrating conic in a work of art allows the picture to be
viewed from the correct viewpoint, which is the point along the ray
perpendicular from the centre of the painting, at a distance such that the
calibrating conic subtends a half-angle of $45^\circ$.  From such a point,
viewers may perceives the painting with accurate perspective as if they were
immersed in the scene.

It is remarkable that we are quite tolerant to the incorrect perspective that
results from viewing a painting from the incorrect position.  This is equally
true in watching motion pictures, where the correct viewpoint may be constantly
changing.  Nevertheless, rapid change of the perspective of an image causes
feelings of discomfort in the viewer. This was the basis for the famous
``Hitchcock zoom'' or ``dolly zoom'' effect in the  film "Vertigo" where it was
used to simulate the effect of dizziness. It is unclear (at least to me) whether
this tolerance of incorrect perspective is a learned, or inate capability.

The position of the horizon, along with the calibrating conic furthermore allows
an estimate of whether the view was taken tilted up or down with respect to the
horizon, and more generally what the relationship is between the viewing angle
and different directions in the image.

\begin{figure}[H]
\centerline{
\includegraphics[height=7.0in]{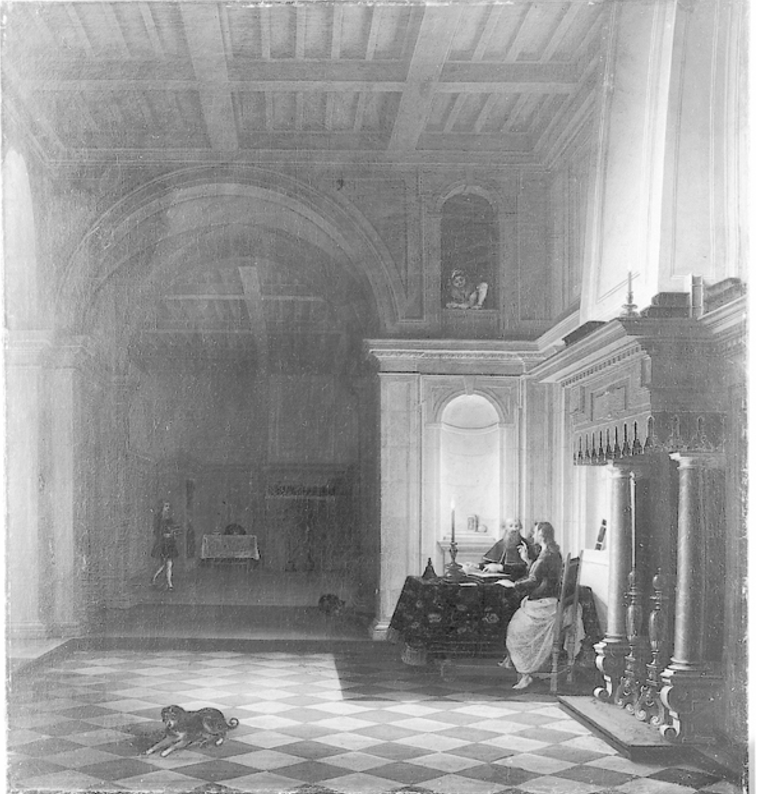}
}
\caption{A painting (Christ Instructing Nicodemus, by Hendrik van Steenwijck, c.\ 1624 ) with a tiled floor.  (Painting kindly identified for me by David Stork.)
}
\label{fig:Nicodemus}
\end{figure}
%
%
\begin{figure}[H]
\centerline{
\includegraphics[width=8.5in]{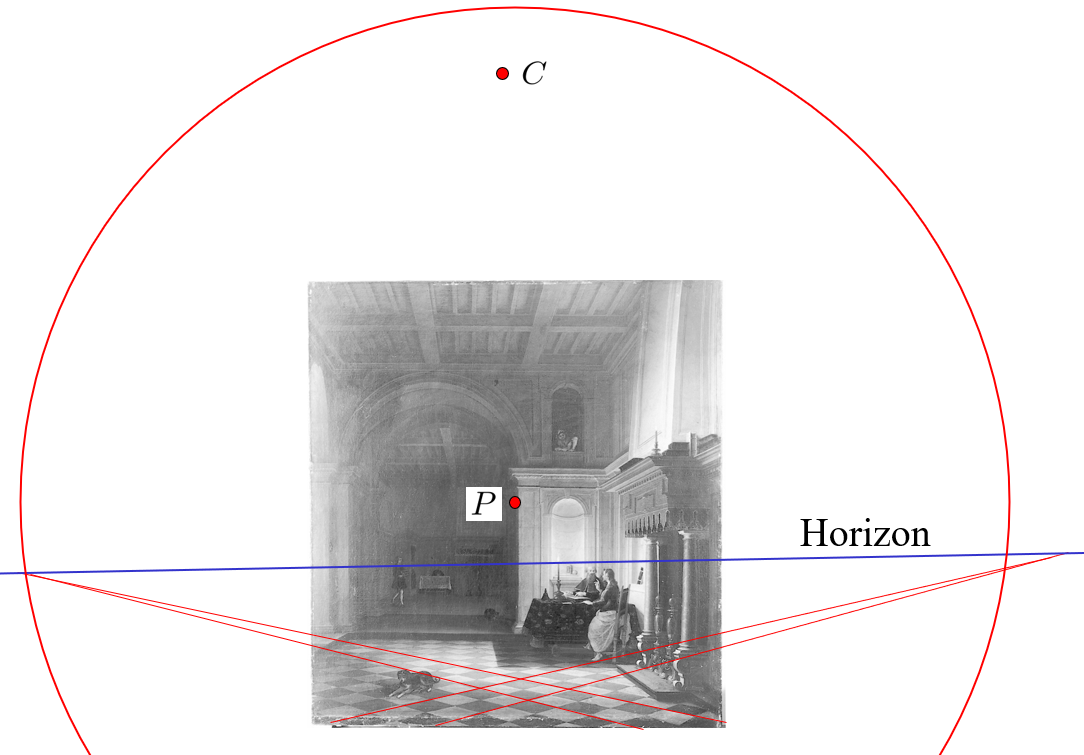}
}
\caption{The calibrating conic can be overlaid on the image.
In the case of art, the calibrating conic can be used to identify the
ideal position from which to view a painting.  The eye should be
place at a position along the perpendicular from the principal point,
at a distance such that the calibrating conic is at $45^\circ$ from the 
line of sight.  From this view point, the viewer experiences the
same perspective effect as the artist.\\
Computation of the calibrating conic identifies this picture as 
representing a relatively narrow field of view.
\\
The field of view may be measured as a half angle of $24.2^\circ$ vertically and $22.5^\circ$ horizontally.
Since the principal point lies above the horizon, the view is in an upward tilted direction.  The
angle of upward tilt is about $6.7^\circ$, as estimated by the method described in \fig{camera-tilt}.
Finally, the view direction is at an angle of about $5.4^\circ$ towards the right with respect to the 
orientation of the room, 
measured by the vanishing point of the floor tiles, the rafters in the roof and the upper edge of the mantle-piece above 
the fireplace.  All of these lines converge at points on the horizon.
\\
(By a rough measure, taking into account a subjective feeling for
the precision of the construction and measurements, an accuracy of around $2\%$ or $3\%$ error
in these estimates seems justified.)
}
\label{fig:art-measurements}
\end{figure}

\clearpage
\section{Odometry -- computing self-rotation in the plane}
In this section it is shown how the use of the conformal point is useful in
self-odometry of a moving vehicle, or robot.  Various elementary facts about
planar structure from motion are cited below. The reader is directed to
\cite{HartleyZisserman:03} or more conveniently to an LLM for clarification, if
needed.

It is assume that a camera is attached rigidly to a moving platform free to move
on a plane.  Thus the only possible motion of the camera is translation in the
plane, and rotation about the vertical axis (perpendicular to the plane). It is
also assumed that the focal length of the camera is fixed, and that pixels are
square. The relevant characteristic of this motion is that under these
circumstances the horizon line of the plane is fixed (as a set, not pointwise)
in the image, and that consequently, the conformal point is also a fixed point.
Therefore, it suffices to compute the conformal point and the horizon line once,
since they remain fixed during the motion.

It is assumed that a set of points on a common plane parallel to the ground,
such as the ground itself or a horizontal roof, are matched from image to image.
 All that is needed is that the two points of each matched point-pair lie on the
same ground-parallel plane. These matches can then be used to determine
pointwise rotations between images.  As a subsequent step, $2$D rotation
averaging can be used to produce a consistent set of rotations for all images.

Once rotations are computed, the translation of the camera is easily computed
linearly.

Thus, the task of computing motion has two important steps, to be elaborated
below:
\begin{enumerate}
\item Compute the horizon and conformal point.
\item Compute pairwise rotations for a pair of images.
\end{enumerate}


\subsection{Computation of the horizon and conformal point}

The method of computing the horizon line in the plane is straight-forward, and
could be accomplished by various methods.  For simplicity we shall assume that
camera is calibrated, and the coordinates in the image are normalized so that
pixels are square, principal point is in the centre of the image, and that the
focal length $f$ is known.

Given a pair of images, one may use matched points to compute a homography
between the two images.  Since the horizon line is invariant under the planar
motion, it can be found as a fixed line under this homography.  This fixed line
is found as an eigenvector of $H\tr$, where $H$ is the computed homography.


\subsection{Computing pairwise rotations}

Now, assuming that the horizon line and conformal point are known. The first
observation is that given two points, denoted $A$ and $B$ in one image, which
are paired with points $A'$ and $B'$ in the other image.  This is sufficient
information to compute the rotation between the two images.  The method is as
follows:
\begin{enumerate}
\item Draw the four points $A$, $B$, $A'$ and $B'$ in the same image plane
(although they come from a pair of images).
\item The angle or rotation of the camera platform is equal to the angle
$\angle(AB, A'B')$ (in the world) between the rays $AB$ and $A'B'$.  This is
readily computed using the conformal point and the known horizon line.
\end{enumerate}
Thus, given a set of point matches between a pair of images, the rotation of the
platform can be computed using any of various robust techniques, such as the
median of a set of rotations computed from two-point matches, or a RANSAC based
method. Since the median can be computed in linear time,  computing the median
of a set of estimated rotation is quick and efficient.

\paragraph{Two-point Ransac. }
As an alternative to computing a median, RANSAC can be used.  The fact that only
two points are required to compute a rotation estimate allows a very efficient
Ransac algorithm.  In outline this is as follows.

\begin{enumerate}
\item Sample pairs of point matches from among a set of matches between views.
\item For each pair of matched points $A \leftrightarrow A'$ and $B
\leftrightarrow B'$ estimate the angle of rotation $\theta = \angle(AB, A'B')$. 
 This notation is meant to indicate the true angle in the world between the
lines in the ground plane, not the angle in the image.
\item Compute the {\em support} for the estimated rotation, and accept the
rotation estimate that has the maximum support.  The support is equal to the
number of points (denoted $D$) whose motion is consistent with a rotation
$\theta$.  This is done as follows. If points $A, B$ and $D$ undergo a rotation
$\theta$ resulting in points $A', B'$ and $D'$, and $\theta = \angle(AB, A'B')$,
then also
\[
\theta = \angle(AD, A'D') = \angle (BD, B'D') ~.
\]
Thus, the match $D \leftrightarrow D'$ supports the estimated rotation $\theta$
if $\angle(AD, A'D')$ and $\angle (BD, B'D')$ equal $\theta$ both equal $\theta$
within a chosen tolerance.
\end{enumerate}

An example of this computation is given in the following figures.

\begin{figure}[H]
\centerline{
\includegraphics[height=4.5in]{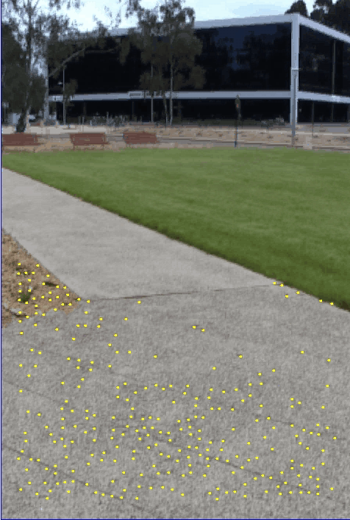}
}
\caption{Points in one image representative of points tracked while the platform moves along the ground plane.
It is necessary only to track points between pairs of images -- complete tracks are not needed. 
From matches, the rotation of the platform is efficiently computed, knowing the conformal point
and horizon of the ground plane, which only need to be computed once.
}
\label{fig:tracked-points}
\end{figure}

\begin{figure}[H]
\centerline{
\includegraphics[height=2.5in]{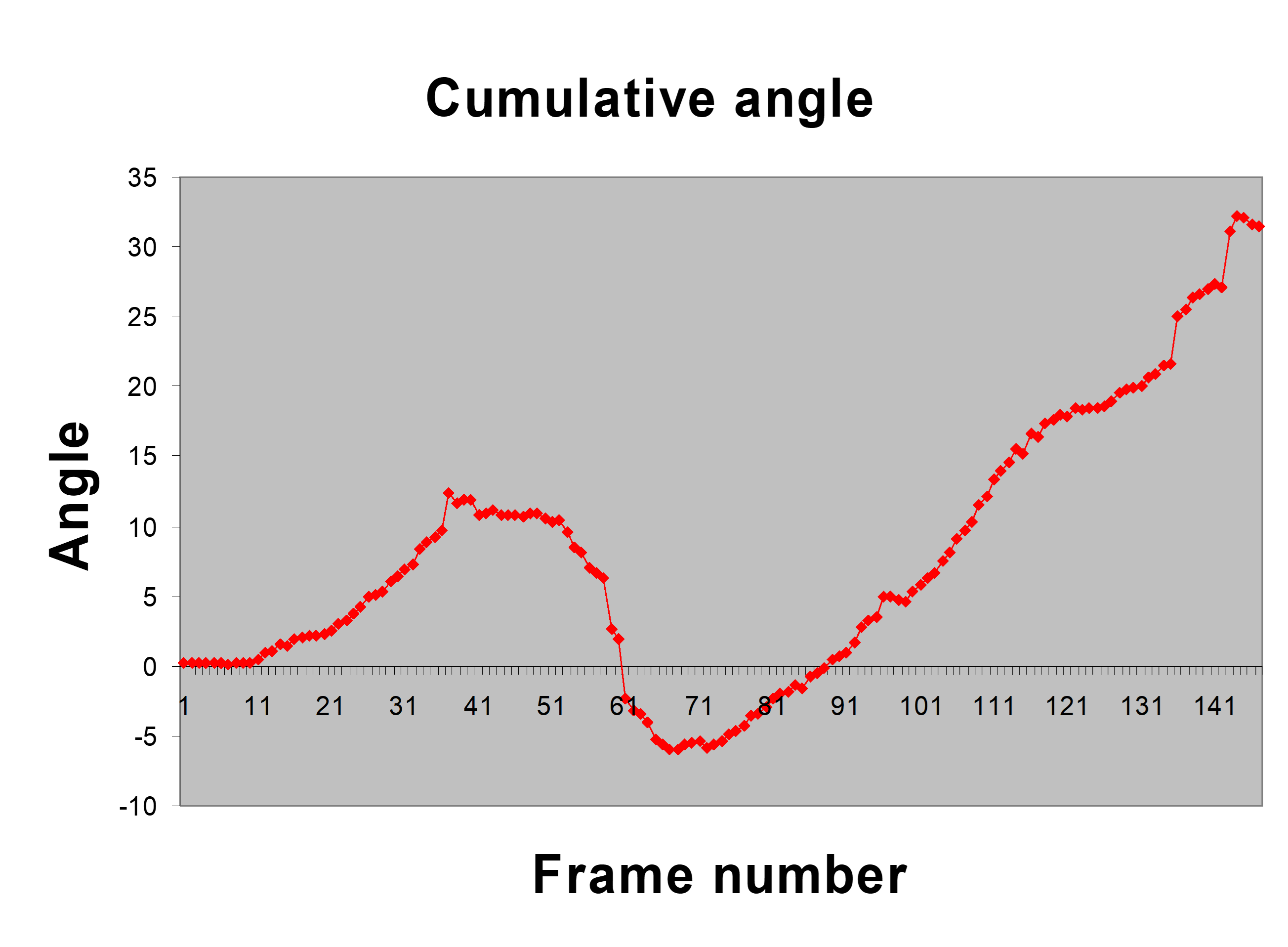}
}
\caption{Estimate of the rotation of the platform through a sequence of $150$ frames.
}
\label{fig:rotation-estimates}
\end{figure}

\bibliographystyle{alpha} \bibliography{ref}

\end{document}

%% file: macros3.tex
\input{local-definitions}
\newcommand{\tr}{\mbox{$^{\top}$}}

\newcommand{\mtr}{\mbox{$^{-\top}$}}

\newcommand{\diag}{\mbox{\rm diag}}

\newcommand{\eq}[1]{(\ref{eq:#1})}

\newcommand{\fig}[1]{fig~\ref{fig:#1}}

\newcommand{\SKIP}[1]{} 

\newcommand{\mbegin} {\left [ \begin{array}}

\newcommand{\mend}   {\end{array} \right ]}

\newcommand{\detbegin} {\left | \begin{array}}
\newcommand{\detend}   {\end{array} \right |}

\newcommand{\vbegin} {\left ( \begin{array}{c}}

\newcommand{\vend} {\end{array}\right )}

\def\squareforqed{\hbox{\rlap{$\sqcap$}$\sqcup$}}

\def\qed{\ifmmode\squareforqed\else{\unskip\nobreak\hfil
	\penalty50\hskip1em\null\nobreak\hfil\squareforqed
	\parfillskip=0pt\finalhyphendemerits=0\endgraf}\fi}

\def\vecs#1{\mathchoice%
	{\mbox{\bf \small $\displaystyle\bf#1$}}
	{\mbox{\bf \small $\textstyle\bf#1$}}
	{\mbox{\bf \small $\scriptstyle\bf#1$}}
	{\mbox{\bf \small $\scriptscriptstyle\bf#1$}}}
\def\vs#1{\protect\vecs #1}

\def\vec#1{\mathchoice%
	{\mbox{\bf $\displaystyle\bf#1$}}
	{\mbox{\bf $\textstyle\bf#1$}}
	{\mbox{\bf $\scriptstyle\bf#1$}}
	{\mbox{\bf $\scriptscriptstyle\bf#1$}}}
\def\v#1{\protect\vec #1}

\newcommand{\showeqnlabel}{
	\hbox to 0pt{\quad\quad\relax\fbox{\scriptsize\rm\eqnlblx}%
	\gdef\eqnlblx{xxxx}}} \newcommand{\eqnlblx}{}

\def\@eqnnum{\rm (\theequation)\showeqnlabel}

\newcommand{\nofig}[1]{\centerline{\bf Figure here}}

\def\mat#1{\mathchoice{\mbox{\bf$\displaystyle\tt#1$}}
	{\mbox{\bf$\textstyle\tt#1$}}
	{\mbox{\bf$\scriptstyle\tt#1$}}
	{\mbox{\bf$\scriptscriptstyle\tt#1$}}}
\def\m#1{\protect\mat #1}